\documentclass[sigconf,nonacm]{acmart}

\usepackage{outlines}
\usepackage{caption}
\usepackage{subcaption}
\usepackage{tabularx}
\usepackage{makecell}
\usepackage[textsize=tiny]{todonotes}
\usepackage{soul}
\usepackage{lipsum}
\usepackage{listings}
\usepackage{xcolor}
\usepackage{amsmath}

\usepackage{adjustbox}
\AtBeginDocument{%
  }

\setcopyright{acmlicensed}
\copyrightyear{2018}
\acmYear{2018}
\acmDOI{XXXXXXX.XXXXXXX}
\acmConference[Conference acronym 'XX]{Make sure to enter the correct
  conference title from your rights confirmation email}{June 03--05,
  2018}{Woodstock, NY}
\acmISBN{978-1-4503-XXXX-X/2018/06}

\begin{document}

\title{The Prompt is Mightier than the Example}

\author{Shengzhe Xu}
\affiliation{%
  \institution{Computer Science \\Virginia Tech}
  \city{Alexandria}
  \state{VA}
  \country{USA}
}

\author{Nikhil Muralidhar}
\affiliation{%
  \institution{Computer Science \\Stevens Institute of Technology}
  \city{Hoboken}
  \state{NJ}
  \country{USA}
}

\author{Naren Ramakrishnan}
\affiliation{%
  \institution{Computer Science \\Virginia Tech}
  \city{Alexandria}
  \state{VA}
  \country{USA}
}

\renewcommand{\shortauthors}{Xu et al.}

\begin{abstract}
Numerous recent prompt optimization approaches like chain-of-thought, tree-of-thought prompting,  have been demonstrated to significantly improve the quality of content generated by large language models (LLMs). In-context learning (ICL), a recent  paradigm where a few representative examples guide content generation has also led to strong and consistent improvements in generation quality of LLM generated content. This idea has been applied to great effect in
synthetic tabular data generation, where LLMs, through effective use of ICL and prompt optimization,
can generate data that approximate samples from complex, heterogeneous distributions based on representative examples.
However, ensuring high-fidelity synthetic data often requires a very large number of ICL examples which may be unavailable or costly to obtain. At the same time, as LLMs get larger and larger, their in-built prior knowledge becomes vast and can potentially substitute for specific data examples. In this paper, we introduce \emph{Knowledge-Guided Prompting} (KGP) as a new knob in prompt optimization and explore the ability of KGP-based prompt optimization to offset the cost of ICL. Specifically,
we explore the question `how many examples can a prompt substitute for?' and explore knowledge-guided prompting (KGP) where domain knowledge, either inferred or available, is explicitly injected into the prompt, reducing dependence on ICL examples. Our experiments systematically explore the trade-off between ICL and KGP, revealing an empirical scaling law that quantifies how quality of generated synthetic data varies with increasing domain knowledge and decreasing example count. We classify prior knowledge into
strong knowledge (e.g.,
symbolic constraints, statistical priors) versus weaker knowledge
(e.g., monotonicity constraints, dependency relationships) and explore
relationships between both forms
and in-context examples. Our results demonstrate that knowledge-guided prompting can be a scalable alternative, or addition, to
in-context examples, unlocking new approaches to synthetic data generation.
\end{abstract}


\maketitle

\section{Introduction}

Synthetic data generation is a key ingredient in many KDD pipelines, e.g., to help overcome privacy limitations~\cite{abay2019privacy,jordon2018pate}, to support machine learning in domains where there are imbalanced classes~\cite{he2009learning}, to enable data augmentation when real data is scarce~\cite{choi2017generating}, and to simulate rare or extreme events that are difficult to capture in real-world datasets~\cite{esteban2017real}. Many powerful ML algorithms, e.g., generative adversarial networks (GANs)~\cite{goodfellow2020generative,hui2023large,wang2020improving,xie2019learning,xu2019modeling} rely on synthetic data generation as a key ingredient to their workflow. 

Recently, large language models (LLMs), especially the latest variants such as GPT-4o and the LLaMA series, have been examined for their potential as structural data regressors or generators.
Most modern LLMs are based on the transformer architecture ~\cite{attention_all_you_need} with parameters ranging from few millions to billions~\cite{chinchilla_scaling_law}, and researchers have developed creative ways to harness LLMs in traditional machine learning and data contexts. For instance, LIFT~\cite{dinh2022lift} transforms table rows of raw numerical data into sentences such as `An Iris plant with sepal length 5.1cm, sepal width 3.5cm...', and employs an LLM to solve traditional machine learning
taasks like classification, regression, and generation.  GReaT~\cite{borisov2022language} fine-tunes an LLM for synthetic data generation and show that even small-scale models such as Distill-GPT~\cite{radford2019language} are capable of synthetic data generation~\cite{borisov2022language}.

While the above works fine-tune an LLM to support data generation, newer variants support synthetic data generation out-of-the-box, i.e., with in-context learning (ICL)~\cite{cllm2024}. 
After prompt engineering paradigms are carefully designed to facilitate in-context learning, just a few example rows in the context window can enable an LLM to generate synthetic data that conforms to the inferred properties of the supplied rows.
However ensuring fidelity to complex, heterogeneous distributions by finding representative examples remains a challenge and requires careful 
experimentation. Just as sampling points on a curve might require more points where there are shifts in behavior (versus regions where there is more normalcy), we will require more ICL examples for some regions versus others to support improved generalization. 

In this paper, we investigate if prompt optimization by injecting prior knowledge into the prompt , can help in synthetic data
generation and, more specifically,
whether it can replicate behavior that previously required an inordinate number of ICL examples. 
We ask the question:
`how many examples can a knowledge-guided prompt substitute for?’ and aim
to capture this tradeoff by
defining knowledge levels and
studying their interplay with 
the number of ICL examples. 
Through our experiments, we show how prompt optimization with explicit domain priors including symbolic, statistical priors, can be infused into prompts to reduce or even eliminate ICL examples.

Our approach is dubbed
knowledge-guided prompting (KGP), where domain knowledge, either inferred or available, is explicitly injected into the prompt (serving as an additional knob for prompt optimization),
reducing the dependence on ICL examples.
This enables new approaches to synthetic data generation than purely data-driven or purely knowledge-driven approaches. 

Our contributions are:

\begin{enumerate}
    \item We propose a new approach for prompt optimization called knowledge-guided prompting (KGP), to improve the quality of structured data generation while at the same time limiting the number of ICL examples required. This approach is especially valuable in scenarios where there is data paucity or we wish to reduce the number of tokens while maintaining synthetic data generation quality.
    \item The KGP approach proposed here systematizes prior knowledge into strong knowledge
(e.g., symbolic constraints, statistical priors) versus weaker knowledge (e.g., monotonicity constraints, dependency relationships), and explores their interplay w.r.t. the number of ICL examples. 
    \item Through numerous  experiments, we demonstrate that our KGP approach yields better generation quality than using purely ICL examples, unlocking new hybrid approaches to synthetic data generation.
    Most importantly, we demonstrate how the KGP framework provides a framework to think about scaling laws that predict the number of examples needed for given levels of prior knowledge.
\end{enumerate}

\begin{figure*}[ht]
  \centering
  \begin{tabular}{c}
    \subfloat[Traditional synthetic tabular data generation pipeline.]{\includegraphics[width=0.85\textwidth]{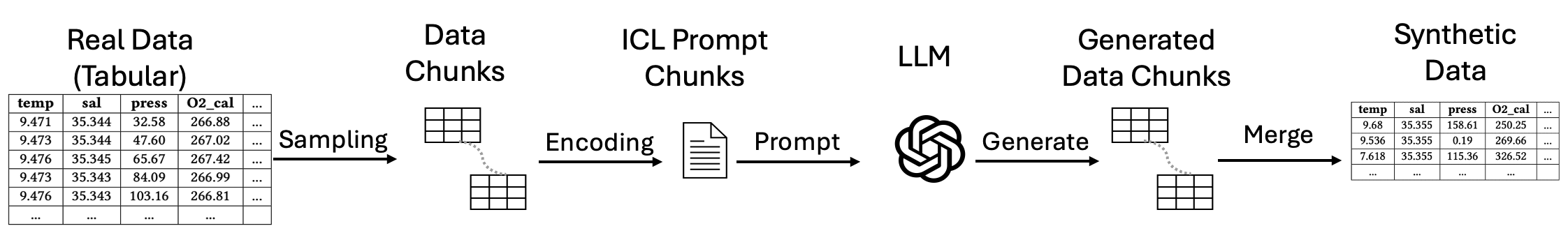}\label{fig:architecture_figure_a}}\\
    \subfloat[Knowledge-guided prompting (KGP) pipeline.]{\includegraphics[width=0.85\textwidth]{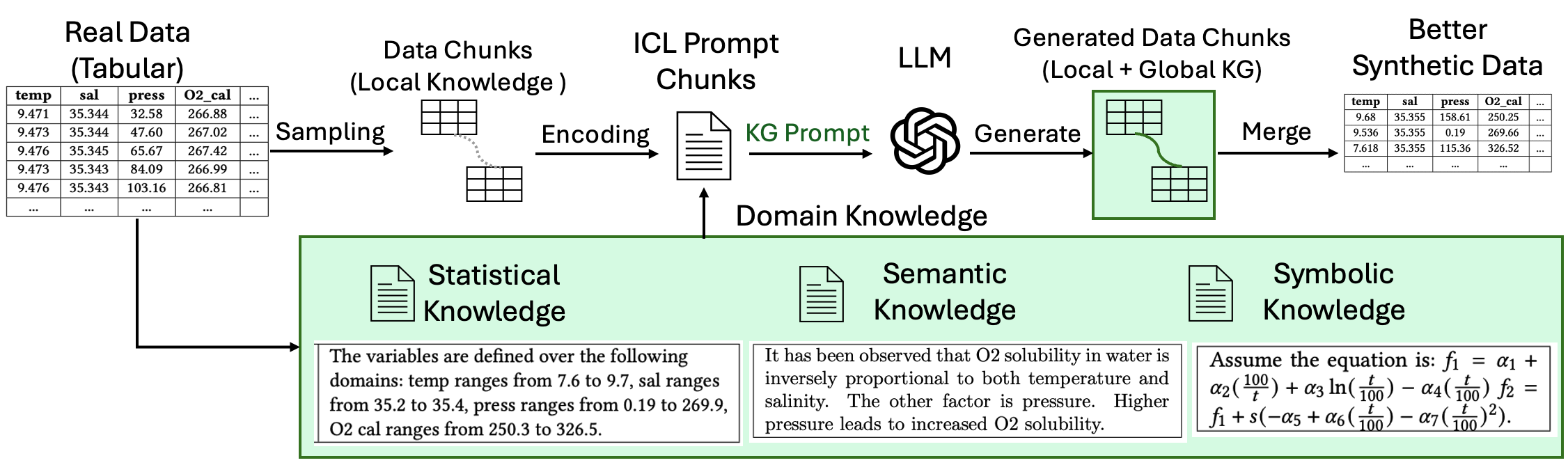}\label{fig:architecture_figure_b}}\\
  \end{tabular}
  \caption{(a) a traditional synthetic tabular data generation pipeline using LLMs encodes sample data as in-context learning examples to drive the generation process. 
  (b) Our prompt optimization approach based on knowledge-guided prompting (KGP), incorporates automatically inferred domain knowledge, providing the LLM-based generator a complementary context in addition to ICL examples. Our experimental findings indicate that such global property conditioning via. KGP leads to a significant improvement in synthetic 
 data generation quality, indicating that KGP can indeed be employed as a useful knob for prompt optimization.}
  \label{fig:architecture_figure}
\end{figure*}

\section{Related Work}
Tabular data synthesis and representation learning for tables have been
extensively studied~\cite{du2021tabularnet, wang2021tuta,zhang2023mixed,margeloiutabebm,fang2024large,SDV,du2024towards,mckenna2019graphical,park2018data,8805442,singha2023tabular}.
For completeness, we survey both
pre-LLM (or non-LLM) and LLM
approaches for synthetic table
generation.

\textbf{Pre-LLM approaches.}
As one of the pre-LLM approaches
to synthetic data generation,
Lei et al.~\cite{xu2019modeling} proposed CTGAN where
rows are independent of each other; a conditional GAN architecture ensures that the dependency between  columns is learned.
Tabsyn~\cite{zhang2023mixed} 
showcased remarkable advancements in joint-distribution learning via a VAE plus diffusion approach, surpassing previous models of similar lineage, in terms of distributional correlation measures and machine learning efficiency.
DoppelGanger~\cite{lin2020using} uses a combination of an RNN and a GAN to incorporate temporal dependencies across rows but this method has been tested in traditional, low-volume settings such as Wikipedia daily visit counts.
For high-volume applications, STAN~\cite{xu2021stan} utilizes a combination of a CNN and Gaussian mixture neural networks
to generate synthetic network traffic data.
GraphDF ~\cite{chen2023graph} is geared toward multi-dimensional time series data.
GOGGLE ~\cite{liu2022goggle} employs a generative modeling method for tabular data by learning relational structures. 

\textbf{LLM approaches.}  
LIFT~\cite{dinh2022lift} 
and GReaT~\cite{borisov2022language} mentioned in the introduction fall in this category.
OmniPred~\cite{omnipred}, provides a framework for training language models as universal
end-to-end regressors over ($x, y$) data from arbitrary formats. Similarly, Treutlein et al. \cite{treutlein2024connecting} exhibit the ability of \textit{inductive out-of-context} reasoning (OOCR) in a regression fine-tuning task of a language model.
(A key difference between these works
and our paper is that in the regression setting, the prompt conditions the output to only predict the target label whereas
we are attempting data conforming to the entire joint distribution at once.)
Recent works~\cite{borisov2022language,xu2024llms,solatorio2023realtabformer,zhang2023generative,zhao2023tabula} have shown the ability to use fine tuning to inject controlled distribution into LLMs, but these approaches are
inflexible and do not leverage prior
LLM's pre-trained knowledge.
Curated LLM~\cite{cllm2024} 
is an approach that
prompts LLMs with specific domain requirements (in English) but this
approach is primarily intended for low-data regimes.
In recent times, instruction-tuned
models have shown great strides in `following instructions' (and some forms of reasoning) but they are still
limited at generating a diversity of
datasets as considered here (some recent efforts~\cite{ren2024towards,dai2022can,hsieh2024ruler,an2024doeseffectivecontextlength}, e.g., BARE~\cite{zhu2025bare}, aim to combine base models with post-training to address this issue).

\textbf{The Prompt vs the Example.} The idea of modeling tradeoffs between prompts and in-context learning (ICL) examples has been studied before~\cite{le2021many}, but primarily in the context of NLP tasks and for a single prompt, not a range of knowledge levels in prompting as studied here. Our work is the first to systematically explore the tradeoff between knowledge and ICL examples for synthetic tabular data generation.
\vspace{-1mm}

\section{Knowledge Guided Tabular Data Generation with LLMs}

Synthetic tabular data generation typically comprises a generation function $\mathcal{G}(\cdot): \mathcal{D}_{\mathrm{train}}\rightarrow \mathcal{D}_{\mathrm{out}}$ where $\mathcal{D}_{\mathrm{train}}$ is the set of samples supplied to $\mathcal{G}$ as input and $\mathcal{D}_{\mathrm{out}}$ is the target data distribution.
The main objective of the tabular data generation task is to generate synthetic data $\mathcal{D}_{\mathrm{syn}}$ conditioned upon $\mathcal{D}_{\mathrm{in}}$ such that $\mathcal{D}_{\mathrm{syn}}\sim \mathcal{D}_{\mathrm{out}}$ i.e., the generated data captures the joint distribution inherent in $\mathcal{D}_{\mathrm{out}}$. 

Recently, LLMs owing to their semantic recognition capabilities as well as pre-trained knowledge, have demonstrated effectiveness in the synthetic tabular data generation task. In the context of LLM based tabular data generation, we can think about the LLM as a few-shot generator where the few-shot nature of the problem arises from the in-context learning (ICL) examples $\mathcal{D}_{\mathrm{train}}$ supplied as input to the LLM-based generator $\mathcal{G}$ as part of the input query $q = [<\mathrm{prompt}>; \mathcal{D}_{\mathrm{train}}]$. The query `q'  comprises the prompt along with $\mathcal{D}_{\mathrm{train}}$ ICL examples.

In the LLM tabular data generation task, owing to the limited effective context windows in LLMs, the input data is chunked into `c' chunks, each a group of $k$ rows  $\mathcal{D}_\mathrm{train} = \{\mathcal{D}^{(1)}_\mathrm{train},\dots,\mathcal{D}^{(c)}_\mathrm{train}\}$ and each chunk is supplied to the LLM as a set of in-context learning (ICL) examples, in addition to a prompt i.e., $q_{i} = [<prompt>;\mathcal{D}^{(i)}_{\mathrm{train}}]$. The result of all the queries  $q_{i} | i=1\dots c$ are merged to form the final generated table $\mathcal{D}_{\mathrm{syn}} = \boldsymbol{\bigcup}^c_{i=1} \mathcal{D}^{(i)}_{\mathrm{syn}}$.
Thus, the LLM, conditioned upon the prompt and ICL examples (i.e., $\mathcal{D}^{(i)}_{\mathrm{train}}$), generates new table rows similar to $\mathcal{D}^{(i)}_{\mathrm{train}}$.  

However, the properties of data in each chunk, $\mathcal{D}^{(i)}_{\mathrm{train}}$ strongly influence the quality of data generated and issues such as \emph{lack of full distributional coverage} and \emph{process noise} may affect the data $\mathcal{D}^{(i)}_{\mathrm{train}}$ thereby carrying over to the generated output chunk $\mathcal{D}^{(i)}_{\mathrm{out}}$, and also create intra-chunk and inter-chunk inconsistensies. To alleviate these adverse effects, we propose knowledge-guided prompting (KGP) as a novel method for prompt optimization by injecting prior domain knowledge about (global) properties prevalent in the ground-truth data distribution in addition to the ICL examples $\mathcal{D}^{(i)}_{\mathrm{train}}$. Essentially, this entails augmenting each query with prior knowledge as follows $q_i = [<\mathrm{knowledge-guided}\, prompt>;\mathcal{D}^{(i)}_\mathrm{train}]$. Prior domain knowledge may occur in many forms and we now detail the various types of domain knowledge and how to inject each into the LLM tabular data generation pipeline via KGP. The full LLM-based tabular data generation pipeline with KGP is depicted in Fig.~\ref{fig:architecture_figure}.

\subsection{Encoded Knowledge Types}
A \textit{knowledge guided prompt (KGP)} accompanying a chunk $\mathcal{D}^{(i)}_\mathrm{train}$ of a dataset $\mathcal{D}_{\mathrm{train}}$, holds for all of $\mathcal{D}_{\mathrm{train}}$. Otherwise stated, KGP encodes global knowledge while a data chunk holds local knowledge.

Prior domain knowledge may appear as symbolic relationships, functional dependencies, semantic descriptions of the data as well as statistical knowledge about the data distribution. The various types of domain knowledge we categorize are illustrated in Table~\ref{tab:knowledge_type}, along with KGP examples of each. More detailed examples in the context of specific datasets investigated in this paper, are included in Table~\ref{tab:kgp_examples}. We specifically focus on three major types of knowledge guidance in this work:
\begin{enumerate}
\item \ul{Symbolic KGP}: In this form of KGP, we assume access to the symbolic (theoretical) relationship governing the (possibly noisy) data generation process.
\item \ul{Semantic KGP}: In this form of KGP, we assume we can encode (partial) knowledge of the data distribution in terms of common prior to take advantage of the semantic recognition capabilities of the LLM.
\item \ul{Statistical KGP}: In this form of KGP, we assume (weak) knowledge about ranges of specific columns in our tabular data.
\end{enumerate}

Fig.~\ref{fig:architecture_figure_b} depicts the proposed KGP pipeline with the various types of domain knowledge considered. Throughout our experimentation, we do not treat all three types of domain knowledge equally, we assume `Statistical KGP' as weak domain knowledge that is the most prevalent, `Semantic KGP' also as weak domain knowledge with relatively lower prevalence than Stastical knowledge and finally we assume `Symbolic KGP' as the strongest as well as the least prevalent type of domain knowledge.

\begin{table}[ht!]
    \centering
    
    \caption{Types of domain knowledge along with examples of how each type can be incorporated into KGP.}
    \begin{tabularx}{\linewidth}{ccX}
        \toprule
        {\bf Type} & {\bf Knowledge} & {\bf Example} \\
        \midrule
        Strong & Symbolic & Equation: $3x^4+ 4x^3 -12x^2 + 2$. \\
        \midrule
        Strong & \makecell[t]{Distribution} & The data follows a specific form of the Bohachevsky function. \\
        \midrule
        Strong & \makecell[t]{Functional\\Dependency} & If Protocol is TCP, then packet size is between 40 to 65,535 bytes.  \\
        \midrule
        Weak & \makecell[t]{Semantic\\Description} & x and y coordinates of points when plotted visually depict a dinosaur. \\
        \midrule
        Weak & \makecell[t]{Statistical\\Knowledge} & The variables are defined over the following domains: temp ranges from 7.6 to 9.7, press ranges from 0.19 to 269.9. \\
        \bottomrule
    \end{tabularx}
    \label{tab:knowledge_type}
\end{table}

\begin{table*}[htpb!]
\small
\centering%
\caption{Example setup of different types of datasets and different levels of knowledge. In practice, the data contains more digits; however, for presentation purposes, we only display up to two to three decimal places.}
\begin{tabularx}{\linewidth}{cXXXXp{3.5cm}}
\toprule
Example Data & W/o KGP & Statistical (Stat.) KGP & Semantic (Sem.) KGP & Symbolic (Sym.) KGP & \makecell[c]{Preview}\\
\midrule
\makecell[tl]{AP Calculus\\(Math)} & x is 2.4278, y is -0.8169. x is 0.2925, y is 1.9153. x is 1.1009, y is -0.1916. [...More] &The variables are defined over the following domains: x ranges from -4.0 to 4.0. & The function f is decreasing if x<=-2, increasing if -2<=x<=0, decreasing if 0<=x<=1, and increasing if x>=1. & Consider the equation: $3x^4+4x^3 -12x^2 +2$. & 
\adjustbox{valign=t,center}{\includegraphics[scale=0.16]{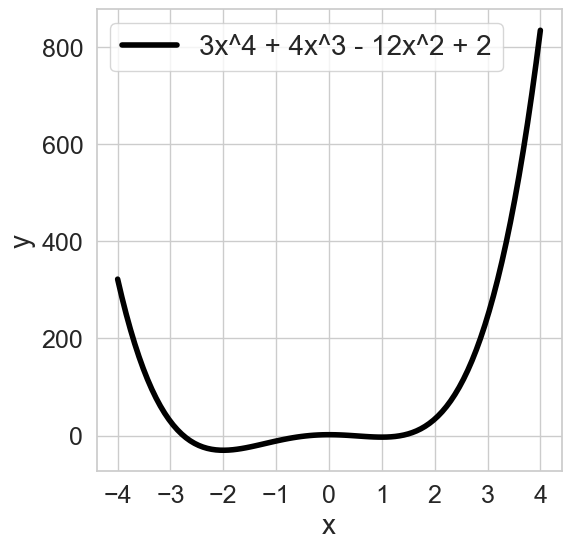}}
\\
\midrule
\makecell[tl]{Datasaurus Dozen\\(Graphical)} & 
x is 55.3846, y is 97.1795. x is 51.5385, y is 96.0256. [...More]
& The range of x is from 31.10686656 to 85.4461864, and the range of y is from 4.57766135 to 97.83761472. &x and y coordinates of points when plotted visually depict a dinosaur.& N/A & \adjustbox{valign=t,center}{\includegraphics[scale=0.16]{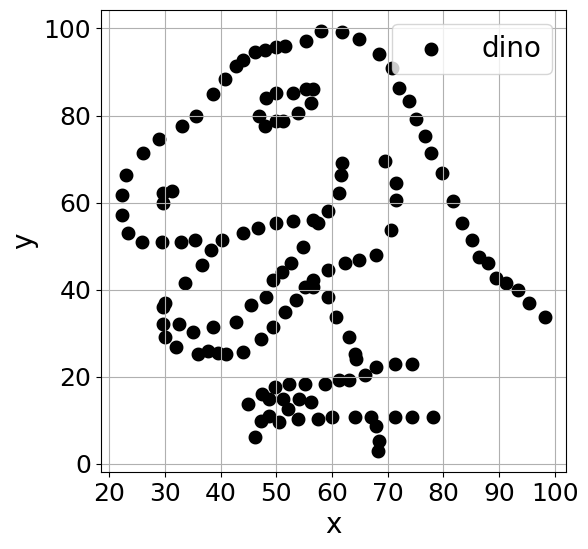}}\\
\midrule
\makecell[tl]{$O_2$ Sensing\\(Real World)} & temp is 9.471, sal is 35.344, press is 32.58, O2 cal is 266.88. temp is 9.473, sal is 35.344, press is 47.60, O2 cal is 267.02. [...More]
   &The variables are defined over the following domains: temp ranges from 7.6 to 9.7, sal ranges from 35.2 to 35.4, press ranges from 0.19 to 269.9, O2 cal ranges from 250.3 to 326.5.& It has been observed that O2 solubility in water is inversely proportional to both temperature and salinity. The other factor is pressure. Higher pressure leads to increased O2 solubility. & N/A & \adjustbox{valign=t,center}{\includegraphics[scale=0.23]{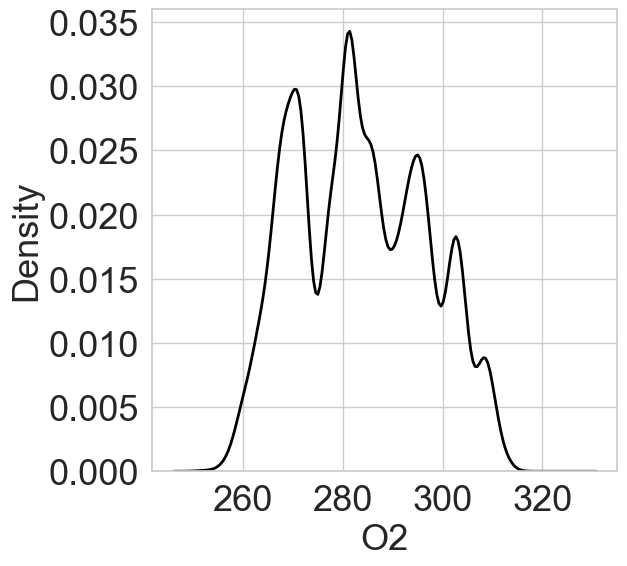}}\\
\bottomrule
\end{tabularx}
\label{tab:kgp_examples}
\end{table*}

\section{Experimental Results}

\begin{figure*}[h]
  \centering
  \includegraphics[width=\linewidth]{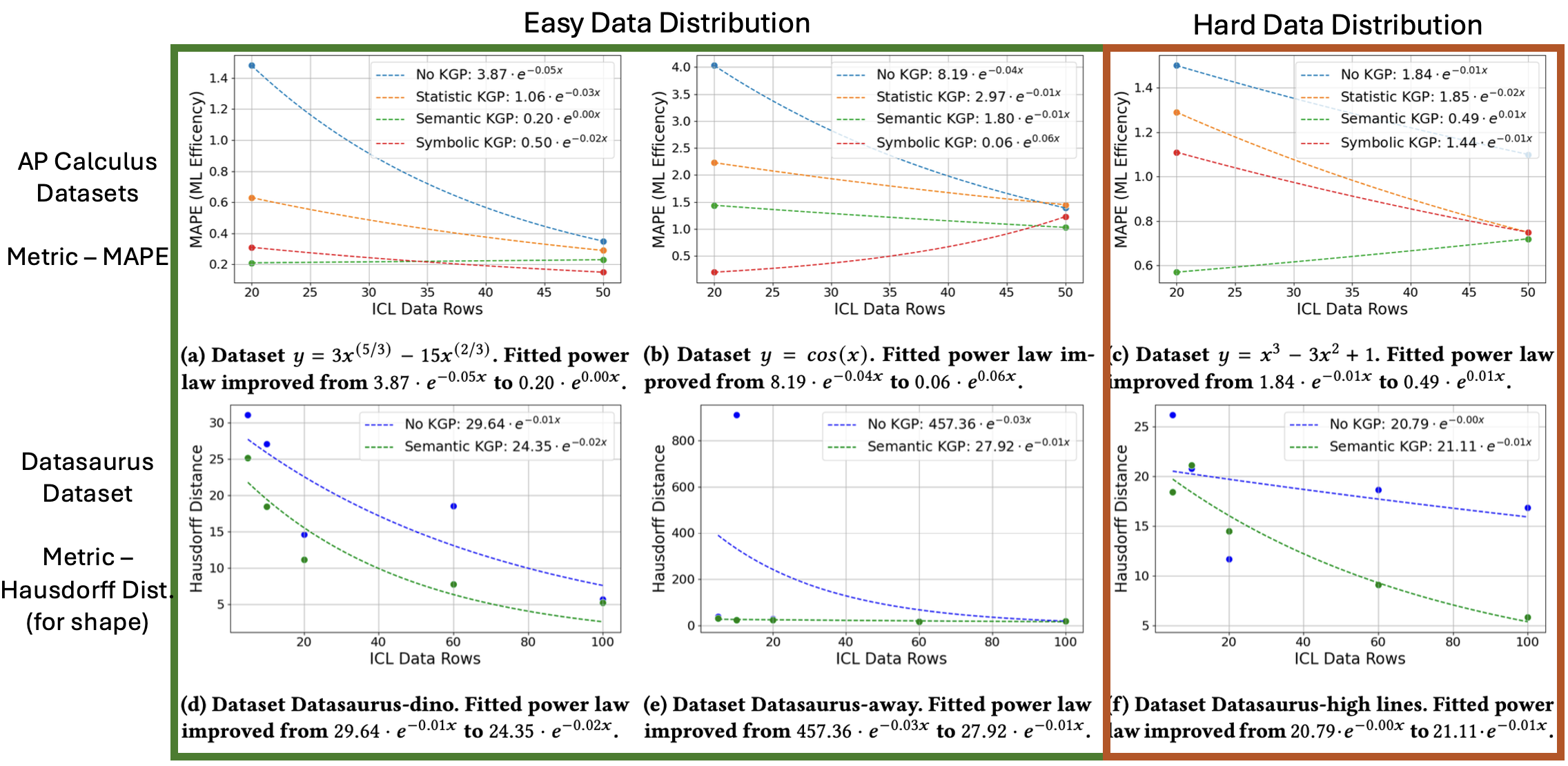}
  \caption{Showcasing the MAPE and Hustoff distance between the synthetic data and the real data. X-axis represents different ICL data sizes. The green curve represents the semantic KGP and the blue curve represents the No-KGP setting. Take (a) for example, by incorporating the visual knowledge phrase ``x and y coordinates of points when plotted visually depict a dinosaur.'' into the prompt, the quality of the generated data improves when the dataset is limited. The quantitative metric Hausdorff Distance decreased from 18.54 to 7.72 indicating a significant improvement when using 60 In-Context Samples.}
  \label{fig:scale_law}
\end{figure*}
In this section, we design experiments on synthetic tabular data generation tasks to investigate the effectiveness of knowledge-guided prompting (KGP) as a novel prompt optimization strategy for LLMs. 
We evaluate our approach using numerous datasets across mathematical, geometric, and real-world applications. Specifically, we wish to investigate the following research questions:

\begin{enumerate}
    \item[\textbf{RQ1}:] What is the trade-off between domain-knowledge and ICL examples? (Section~\ref{sec:rq1})
    \item[\textbf{RQ2}:] Can domain-knowledge alleviate effects of  poor data coverage or help with (out-of-domain) OOD generalization? (Section~\ref{sec:rq2})
    \item[\textbf{RQ3}:] Which type of knowledge injection is the most effective? (Section~\ref{sec:rq3})
    \item[\textbf{RQ4}:] How does KGP affect the quality of the synthetic data generated? (Section~\ref{sec:rq4})  
    \item[\textbf{RQ5}:] (Case Study) Can we characterize the effectiveness of KGP in a real-world cyber-physical scenario? (Section~\ref{sec:case1})
\end{enumerate}

\subsection{Setup \& KGP Scope}
\textbf{Datasets.} We have adopted real datasets across three application domains. We manually extracted datasets from the AP Calculus textbook (specifically, Section 4~\cite{anton2011calculus}), featuring variations of equations and descriptions of function characteristics,
Thirteen datasets from the Datasaurus Dozen exhibiting distinct visual characteristics~\cite{matejka2017same}, and an $O_2$ sensing dataset from real-world applications related to cyber-physical systems~\footnote{https://www.bco-dmo.org/dataset/3426}.

\textbf{Baselines.} We aim to investigate the potential of in-context prompting techniques utilizing large language models, and in this experiment the flagship OpenAI model GPT-4o is utilized as the foundation model.
In accordance with the system outlined in the previous section, three levels of knowledge-guidance prompts will be introduced and analyzed in an ablation study: Statistical KGP, Semantic KGP, and Symbolic KGP.

\textbf{KGP Scope.} It is important to clarify that for the purpose of this
evaluation, we treat the levels of knowledge as a set of
concentric circles. In other words,
``Semantic KGP'' denotes the \textbf{combination of Statistical and Semantic KGP}. Similarly, ``Symbolic KGP'' includes \textbf{all three forms of knowledge}. Notably, Without knowledge implies no guidance from knowledge is utilized, serving as the traditional baseline for synthetic data generation.

\textbf{Metrics.} 
The traditional metrics for synthetic data from the tabular data generation community are utilized, including machine learning utility (MLU), negative log likelihood (NLL), KL divergence, and distance to the closest record (DCR).
Additionally, for datasets containing ground-truth symbolic equations, we employ mean square error (MSE) as the primary metric for assessing data record validity. For datasets characterized by shape-focused distributions, we employ Hausdorff distance to assess the similarity of the shapes.

\subsection{RQ1: What is the trade-off between domain-knowledge and ICL examples?}
\label{sec:rq1}

Encoding structural data within a text sentence for LLM utilization incurs a substantial token load. This underscores the rationale for saving example data tokens to maintain performance expectations or to enhance performance in scenarios where example data is insufficient.

Although LLM developers continue to explore the upper limits of context windows for both input and output, their efforts remain insufficient in the domain of synthetic structural data generation.
Therefore, it is important to consider strategies for conserving tokens utilized by in-context data samples, as well as identifying ways to assist a language model when a user's example data is limited. Here are two common cases:

\par\noindent
\textbf{Simple Data Distributions}. When the joint distribution of the data is easy to model, using KGP will decrease data requirements and thereby reduce token demand, leading to financial and time saving. Figure~\ref{fig:scale_law}a,b,c and e investigate the impact of KGP on data generation when the target data follows a simple joint distribution. Specifically, Figure~\ref{fig:scale_law}a, b have been evaluated on datasets from the AP calculus~\cite{anton2011calculus} data corpus, where 4 KGP variants have been evaluated enabling us to investigate the full range of knowledge-guidance granularity (i.e., statistical, semantic and symbolic knowledge guidance). For each KGP variant, one context with 20 ICL examples and another with 50 ICL examples has been evaluated. 
The plots in Figure~\ref{fig:scale_law}a,b both clearly demonstrate the benefit of KGP over the `No KGP' variant in low data (i.e., 20 ICL examples) scenarios. 
\par\noindent
\textit{\ul{Finding}}: \textit{Semantic KGP, Symbolic KGP require 40\% fewer ICL examples to achieve the same generation quality as a variant without KGP.}

\begin{figure*}[t!]
  \centering
  \resizebox{0.8\textwidth}{!}{%
  \begin{tabular}{c|ccc}
    \subfloat[ICL Data \& Full Scope]{\includegraphics[width=0.22\textwidth]{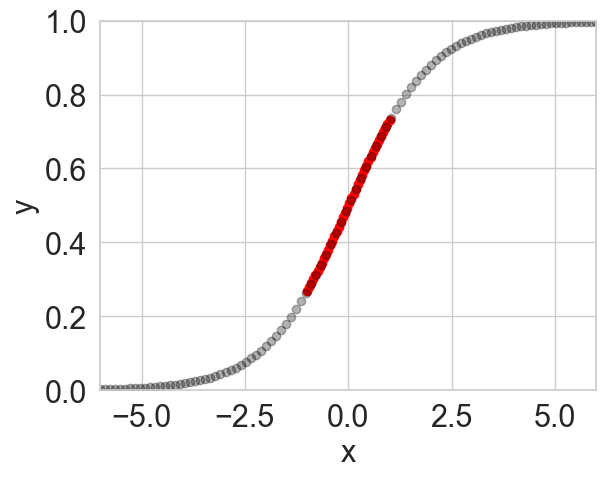}} & \subfloat[W/o KGP]{\includegraphics[width=0.22\textwidth]{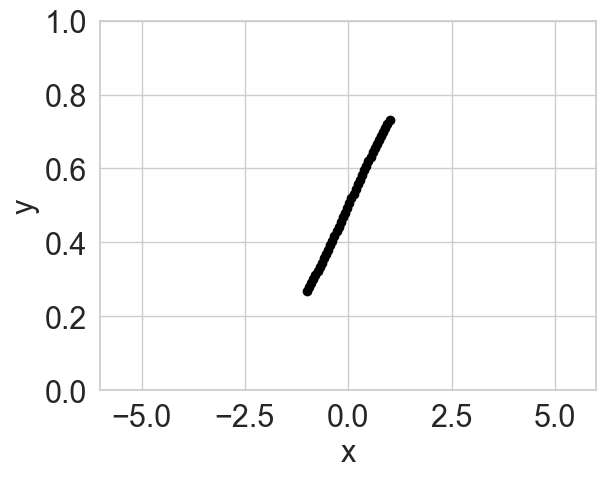}}&\subfloat[Statistical KGP]{\includegraphics[width=0.22\textwidth]{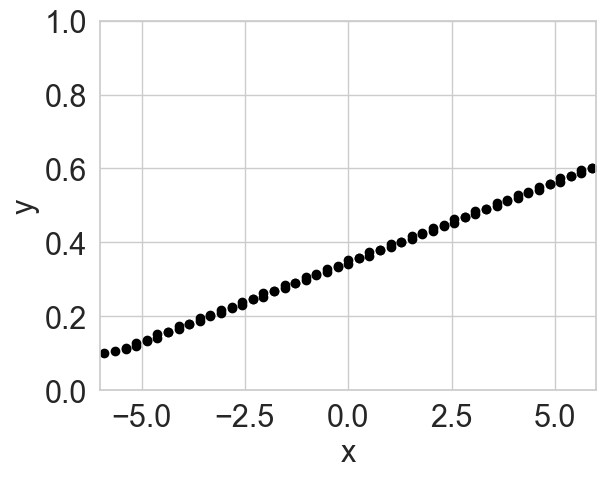}\label{fig:ood_vis_c}}& \subfloat[Semantic KGP]{\includegraphics[width=0.22\textwidth]{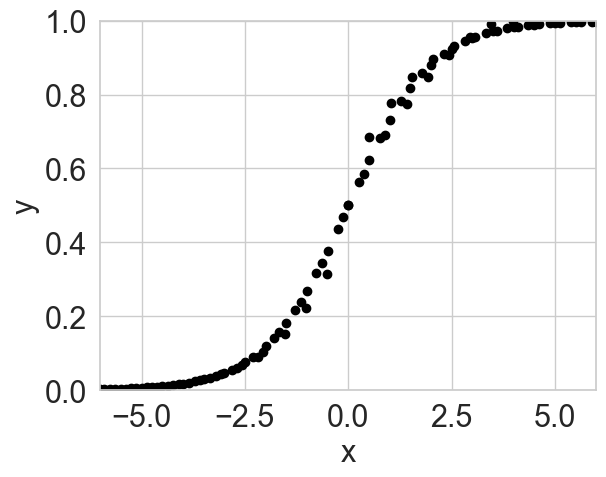}} \\ 
     \subfloat[ICL Data \& Full Scope]{\includegraphics[width=0.22\textwidth]{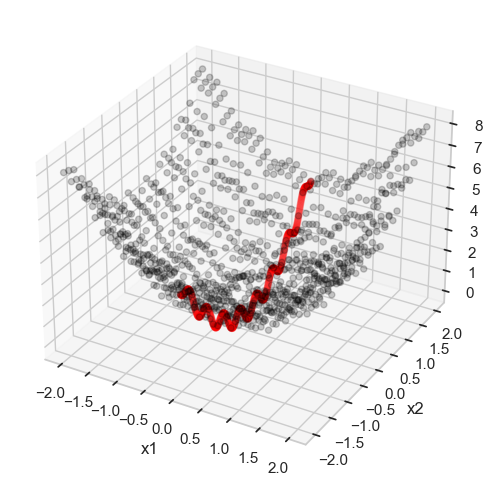}} & \subfloat[W/o KGP]{\includegraphics[width=0.22\textwidth]{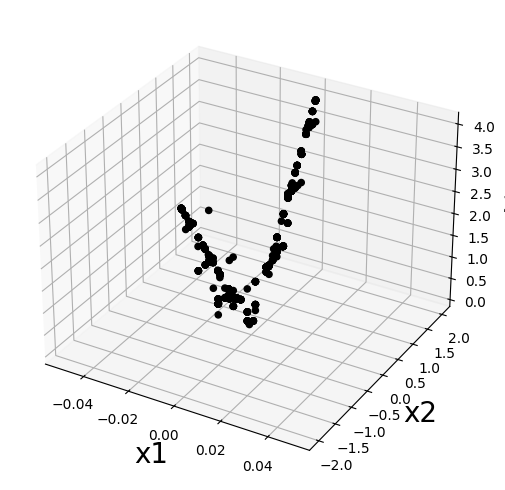}}&\subfloat[Statistical KGP]{\includegraphics[width=0.22\textwidth]{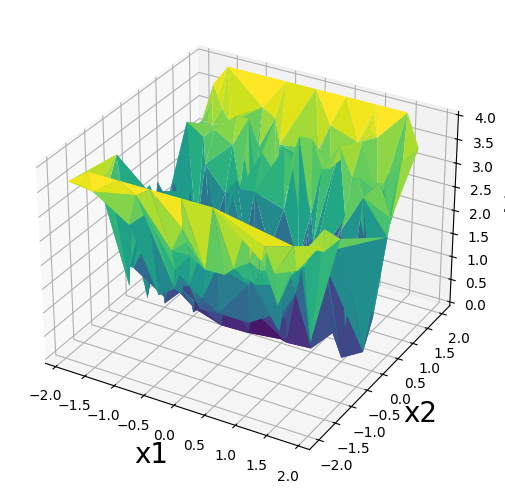}\label{fig:ood_vis_g}}& \subfloat[Semantic KGP]{\includegraphics[width=0.22\textwidth]{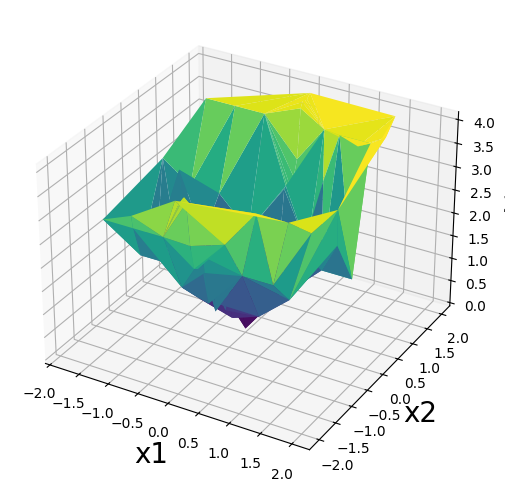}} \\ 
  \end{tabular}
  }
  \caption{Visualization of out-of-distribution (OOD) generation, featuring two mathematical functions: Sigmoid and Bohachevsky. In the ICL Real Data figure (a) \& (e), the \textcolor{red}{red} data points represent the observed field, whereas the \textcolor{black!50}{grey} data points indicate the complete ground truth field. Figure (b)-(d), (f)-(h) showcase the generated synthetic data under corresponding KGP settings.}
  \label{fig:ood_vis}
\end{figure*}

Further, a similar experiment is carried out on the Datasaurus corpus~\cite{matejka2017same} employing the popular Hausdorff distance metric to test data generation quality. In this scenario, we compare the `No KGP' variant with the `Semantic` KGP variant. The two variants are each evaluated in two ICL contexts namely, one with 10 and another with a 100 random ICL data points. The goal is once again to evaluate how the KGP affects tabular data generation quality and its utility in low-data scenarios.

Figure~\ref{fig:scale_law}d evaluated in the context of the Dino dataset from the Datasaurus corpus, illustrates that `Semantic KGP' achieves equivalent generation quality as `No-KGP' with a 40\% reduction in the number of ICL examples. Figure~\ref{fig:scale_law}e is a similar comparison performed on the \emph{Away dataset} from the Datasaurus corpus and demonstrates an even higher reduction (ie., 90\%) in ICL examples in the Semantic KGP context to achieve thee same generation quality as the No KGP context.

\par\noindent
\textit{\ul{Finding}}: \textit{Overall, KGP improves synthetic-data generation quality with a 40\% - 90\% reduction in ICL examples while achieving the same generation quality as a variant without KGP, even in for simple data distributions.}

\par\noindent
\textbf{Complex Data Distributions.} In Figure~\ref{fig:scale_law}c and ~\ref{fig:scale_law}f, we evaluate datasets from the AP Calculus and Datasaurus corpora respectively except here, we consider datasets where the data exhibits a more complex (i.e., harder to model) joint distribution. 
Figure~\ref{fig:scale_law}f illustrates an example of modeling a relatively difficult joint distribution (i.e., \emph{High Lines} dataset) which is difficult owing to the data being distributed in  disparate statistical modes. Here, we notice that despite being conditioned on 100 in-context samples, even a state-of-the-art LLM like GPT-4o alone (i.e.,  (with No KGP) does not generate a valid synthetic joint distribution (as evidenced by high Hausdorff distance of the blue line even at ICL 100). However, by simply injecting a semantic KGP statement such as `x and y are visually looking like high lines', the model can not only significantly improve the quality of generated data but achieves the same generation quality as `No KGP' with 80\% fewer ICL examples.

Figure~\ref{fig:scale_law} represents a cubic polynomial function, also hard to model in a purely data-driven manner. We notice that incorporating any form of KGP (statistical, semantic or Symbolic) leads to a significant reduction in data generation error and a 50\% reduction in ICL examples to achieve the same generation quality as the `No KGP' variant.
\par\noindent
\textit{\ul{Overall Finding}}: \textit{KGP results in a significant reduction in ICL examples for synthetic tabular data generation, both in contexts where the data follows an easy and a hard joint distribution. Specifically, knowledge guidance leads to a  40\%-90\% reduction in ICL examples in the easy data context and between 50\%-80\% reduction in the hard data context.}

\subsection{RQ2: Can domain-knowledge alleviate effects of  poor data coverage or help with OOD generalization?}
\label{sec:rq2}

Table~\ref{tab:ood_mse} showcases the capability of  generating previously unobserved structural data on the basis of the `Statistical' and `Semantic' KGP provided, while Figure~\ref{fig:ood_vis} illustrates the visual representation. 
When calculating the MSE of the uncovered field of the `No KGP` variant, noise will be examined. `Statistical' KGP and `Semantic' KGP will generate data to cover those region utilizing the injected domain knowledge.
The mean squared error (MSE) of the sigmoid function can be significantly reduced by 98\%.
Furthermore, with respect to the Bohachevsky function, the absolute value of the error decreased from 1.62 to 0.44 due to its higher dimensionality and increased complexity. 
In the realm of complex functions, delving into unfamiliar areas demands increased caution, as the LLM generator may mistakenly treat unknown fields as similar to known ones, as illustrated in Figure~\ref{fig:ood_vis_c} and~\ref{fig:ood_vis_g}.

\begin{table}[htb]
  \centering
  \small
  \caption{MSE for OOD generalization.}
  \begin{tabular}{c|c|cc|c}
    \toprule
    Math Function & W/o KGP & \makecell{Statistical\\ KGP} & \makecell{Semantic\\KGP} & \makecell{MSE\\Impr.}\\
    \midrule
    Sigmoid (2d)   & 0.11   & 0.09  & \textbf{0.002} & $\downarrow$ \textbf{98\%} \\
    Bohachevsky (3d)   & 1.62   & 2.23 & \textbf{0.44} & $\downarrow$ \textbf{73\%}  \\
    \bottomrule
  \end{tabular}
  \label{tab:ood_mse}
\end{table}

\par\noindent
\textit{\ul{Overall Finding}}: \textit{
With the support of domain knowledge, KGP is capable of generating out-of-distribution (OOD) data and augmenting datasets that suffer from poor coverage or missing values. The data generated through `Statistical' plus `Semantic' KGP exhibits an error rate that is 78\% to 90\% lower compared to the plain `No KGP' method when exploring to unknown data feild.}

\subsection{RQ3: Which type of knowledge injection is the most effective? }
\label{sec:rq3}

Table~\ref{tab:types_effect} shows that the injection of `Statistical' KGP and `Semantic' KGP generally leads to consistent and improved data quality.
Utilizing the complete equation, notably `Symbolic' KGP, does not always produce beneficial outcomes due to the limited grasp of complex mathematics.

\par\noindent
\textit{\ul{Overall Finding}}: \textit{The integration of statistical KGP and semantic KGP in data generation involving various mathematical function relationships can produce a consistent improvement in quality (i.e., a reduction in MSE), ranging from 35\% to 70\%, without occasionally causing negative error.}

\begin{table}[htbp!]
\small
\centering%
\caption{Mean Squared Error at the 50-ICL setting with various levels of KGP. Arrows indicate the trend of the effect (MSE) as higher levels of KGP (i.e., more granular knowledge-guidance rules) are injected.}
\begin{tabularx}{\linewidth}{p{2.8cm}XXXX}
\toprule
Math Function & W/o KGP & Statistical KGP & Semantic KGP & Symbolic KGP \\
\midrule
$y = 3x^{(5/3)} - 15x^{(2/3)}$ &0.35 & ($\downarrow$)0.29 & ($\downarrow$)0.23  & ($\downarrow$)\textbf{0.15}\\
\midrule
$y = x^3-3x^2+1$ &1.10& ($\downarrow$)0.75& ($\sim$)\textbf{0.72}& ($\uparrow$)0.75\\
\midrule
$y = 2x^3-15x^2+36x$ &0.02& ($\sim$)0.02& ($\sim$)0.02& ($\downarrow$)\textbf{0.01}\\
\midrule
$y = x+2sin(x)$ &0.40& ($\downarrow$)0.14& ($\sim$)\textbf{0.12}& ($\uparrow$)0.57\\
\bottomrule
\end{tabularx}
\label{tab:types_effect}
\end{table}

\subsection{RQ4: How does KGP affect the quality of generated synthetic data?}
\label{sec:rq4}
In addition to the savings on the number of ICL examples, it is equally crucial for KGP based generation pipelines to ensure high-quality synthetic data. To investigate this, we quantitatively evaluate the synthetic data quality with well accepted metrics: low-order statistics (Sec~\ref{sec:rq4_low_order}), machine learning utility (MLU) (Sec~\ref{sec:rq4_mle}), and closest distance to record (DCR) (Sec~\ref{sec:rq4_dcr}).

\subsubsection{How close is synthetic data to the full distribution joint (low-order statistics)?}
\label{sec:rq4_low_order}

Table~\ref{tab:low_order} provides a quantitative assessment of the performance of the synthetic table. The negative log likelihood (NLL) quantifies the resemblance between synthetic data and real data. The low KL-divergence simultaneously guarantees the mode diversity of the synthetic data.

\begin{table}[h]
    \small
    \centering
    \caption{Low-order statistics, evaluated by negative log likelihood (NLL) and KL-divergence. Both of the metrics are smaller the better.}
    \begin{tabularx}{\linewidth}{Xccc}
        \toprule
        \multicolumn{4}{c}{Negative Log Likelihood (NLL) ($\downarrow$)~} \\
        \midrule
        Dataset & W/o KGP & \makecell{Statistical\\KGP} & \makecell{Semantic\\KGP} \\
        \midrule
        AP Calculus & 4.75$\pm$0.90 & \textbf{4.73}$\pm$\textbf{0.73}& 4.91$\pm$1.14 \\
        \midrule
        Datasaurus Dozen & 8.99$\pm$0.18 & 9.09$\pm$0.25 & \textbf{8.88}$\pm$\textbf{0.04} \\
        \midrule
         $O_2$ Sensing & 31.18 & 11.62 & \textbf{8.54}  \\
        \bottomrule
        \toprule
        \multicolumn{4}{c}{KL-Divergence ($\downarrow$)~} \\
        \midrule
        Dataset & W/o KGP & \makecell{Statistical\\KGP} & \makecell{Semantic\\KGP} \\
        \midrule
        AP Calculus & \textbf{0.02}$\pm$\textbf{0.02} & 0.04$\pm$0.04 & 0.07$\pm$0.09 \\
        \midrule
        Datasaurus Dozen & 0.06$\pm$0.07 &  0.10$\pm$0.06 & \textbf{0.01}$\pm$\textbf{0.01} \\
        \midrule
         $O_2$ Sensing & 4.43 & 0.70& \textbf{0.20}  \\
        \bottomrule
    \end{tabularx}
    \label{tab:low_order}
\end{table}

\begin{figure}[htb!]
  \centering
  \begin{subfigure}[b]{0.35\textwidth}
    \centering
    \includegraphics[width=\textwidth]{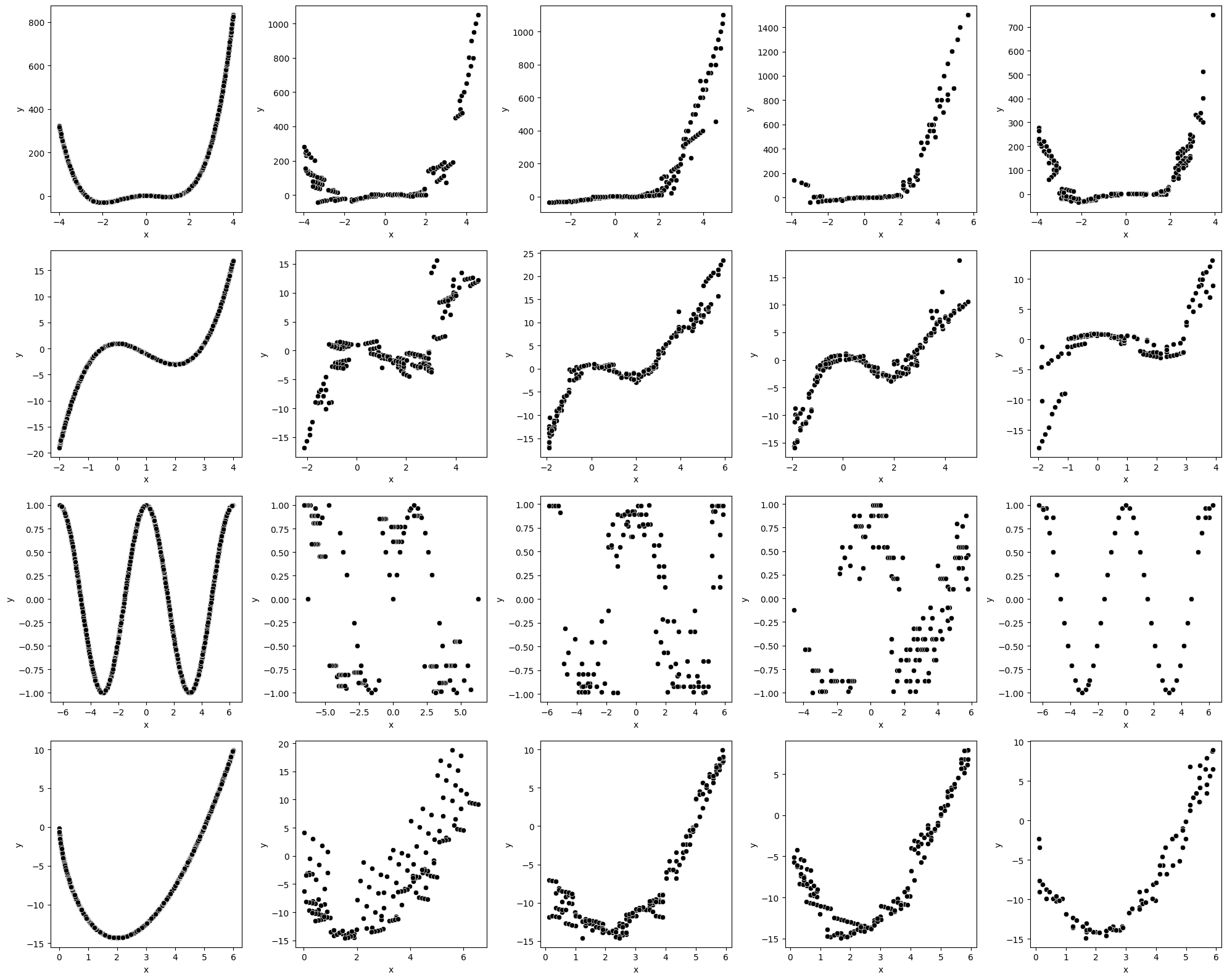}
    \caption{ICL-20 for four functions, one per row: ($3x^4+4x^3-12x^2 +2$); ($x^3 - 3x^2 + 1$); ($cos(x)$); ($3x^{(5/3)} - 15x^{(2/3)}$).}
    \label{fig:subfigure1}
  \end{subfigure}
  \hfill
  \begin{subfigure}[b]{0.35\textwidth}
    \centering
    \includegraphics[width=\textwidth]{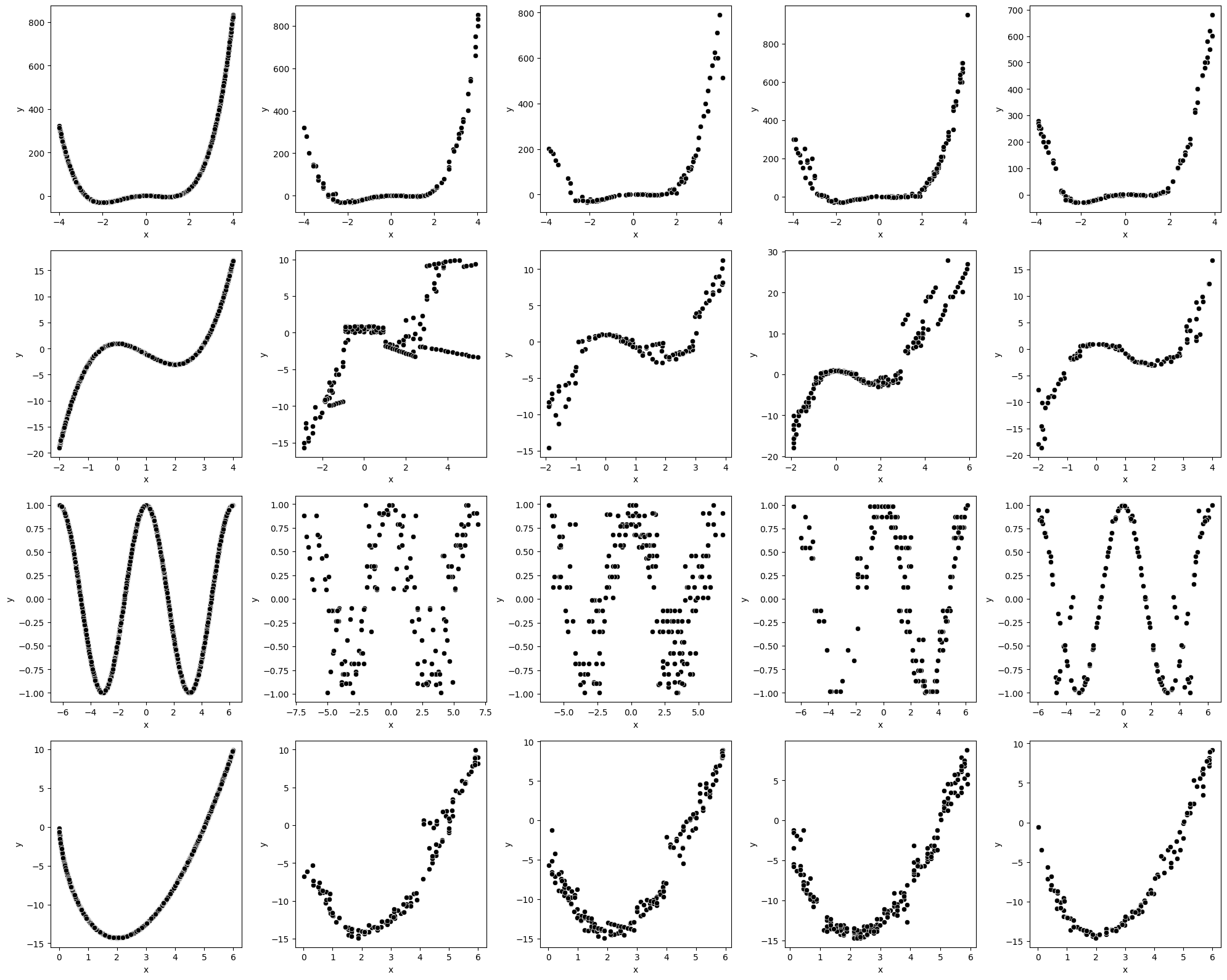}
    \caption{ICL-50 for four functions, one per row: ($3x^4+4x^3-12x^2 +2$); ($x^3 - 3x^2 + 1$); ($cos(x)$); ($3x^{(5/3)} - 15x^{(2/3)}$)}
    \label{fig:subfigure2}
  \end{subfigure}
  \caption{Diversity of modes in synthetic data. Five columns from left to the right are real data, No KGP, statistical KGP, semantic KGP, and symbolic KGP.}
  \label{fig:mode_diversity_ap}
\end{figure}

\begin{figure}[h]
  \centering
  \begin{tabular}{c}
    \subfloat[Good Semantic Knowledge: `Dinausour', `x shape', `star'.]{\includegraphics[width=0.3\textwidth]{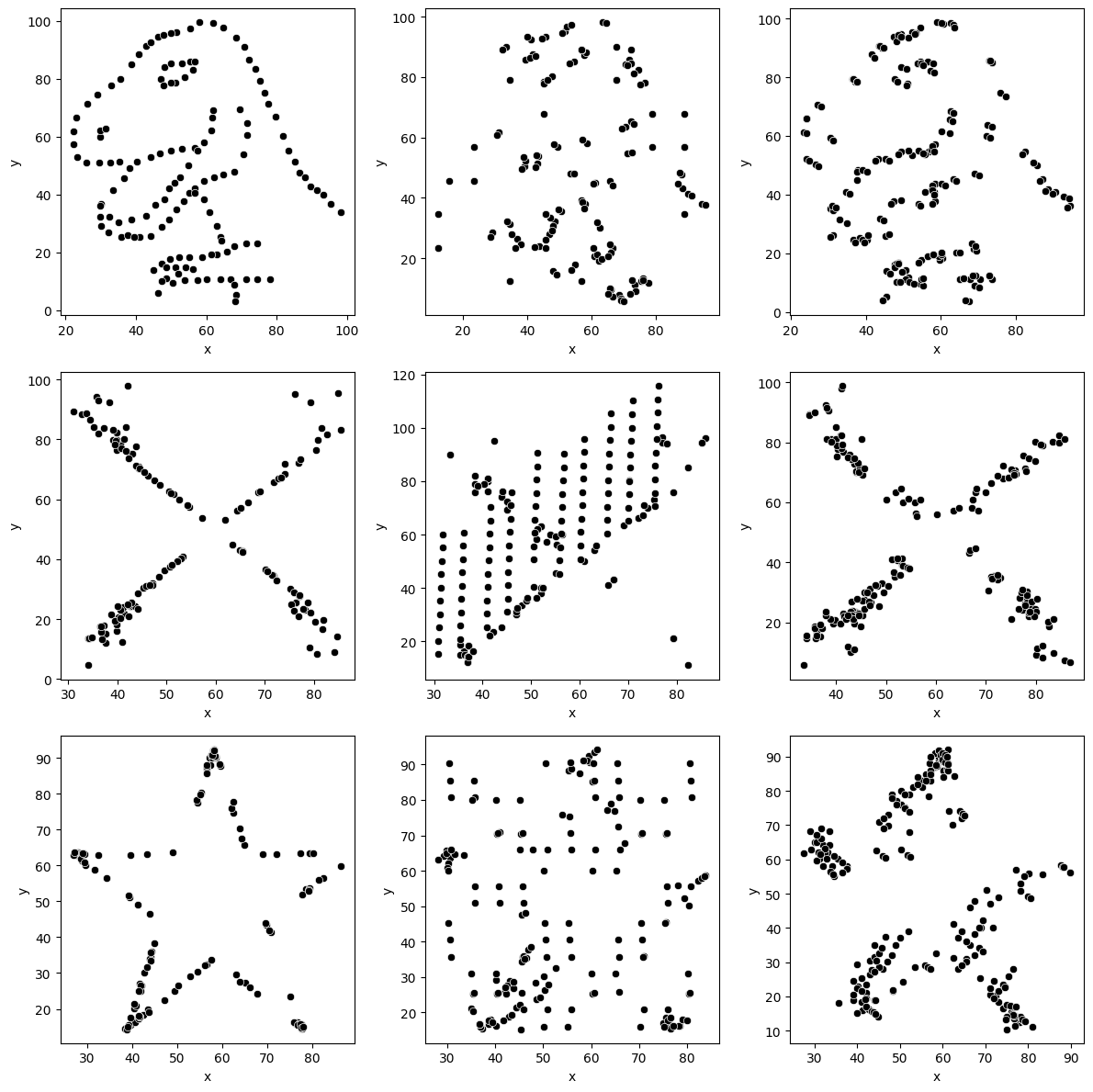}} \\ 
     \subfloat[Misleading Semantic Knowledge: `bullseye', `slant up', `wide lines'.]{\includegraphics[width=0.3\textwidth]{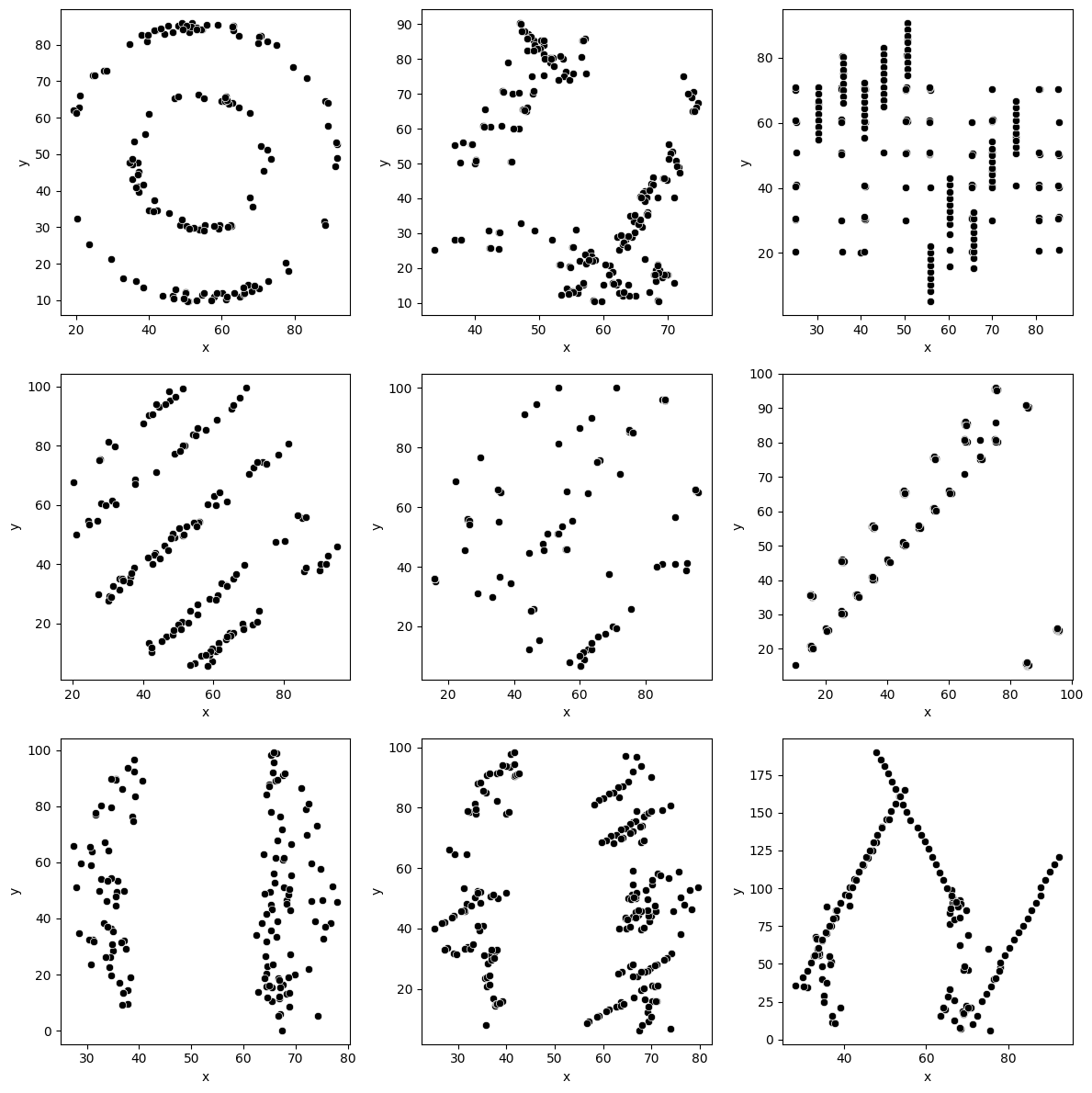}} \\ 
  \end{tabular}
  \caption{Diversity of modes in synthetic data. Three columns from left to the right represent real data, No KGP, Semantic KGP.}
  \label{fig:mode_diversity_dino}
\end{figure}

Figure~\ref{fig:mode_diversity_ap} compares real and synthetic data, highlighting shape trends. Analyzing columns two (no KGP) to five (Symbolic KGP) shows that better knowledge guidance yields more realistic results within the same sample. Figure (a) represents the ICL20 conditions, while Figure (b) illustrates the ICL50 conditions. 
Comparison of subfigures (a) and (b) in Figure~\ref{fig:mode_diversity_dino} shows that the Semantic KGP form in graphs requires a more detailed context in English. Otherwise, the Semantic KGP can similarly contaminate synthetic data, akin to a reverse de-noising process.

\subsubsection{How does distance to the closest record change when knowledge is incorporated?}
\label{sec:rq4_dcr}
Table~\ref{tab:dcr_table} shows improved row similarity without adding a new leak record.

\begin{table}[h]
    \small
    \centering
    \caption{Distance to the closest record: A lower distance yields a better record in terms of validity; however, the occurrence of a zero value, which indicates a leak of raw data, is unacceptable.}
    \begin{tabularx}{\linewidth}{X|c|cc}
        \toprule
        \multicolumn{4}{c}{Distance to the closest record. ($\downarrow$)~} \\
        \midrule
        Dataset & W/o KGP & \makecell{Statistical\\KGP} & \makecell{Semantic\\KGP} \\
        \midrule
        AP Calculus & \textbf{0} & 0 & 0 \\
        \midrule
        Datasaurus Dozen & 0.21$\pm$0.20 & 0.41$\pm$0.24 & \textbf{0.12}$\pm$\textbf{0.09} \\
        \midrule
         $O_2$ Sensing & 0.57 & 0.46 & \textbf{0.38} \\
        \bottomrule
    \end{tabularx}
    \label{tab:dcr_table}
\end{table}

\subsubsection{Can we use synthetic data in ML pipelines?}
\label{sec:rq4_mle}

Table~\ref{tab:mle} illustrates the performance of machine learning using synthetic data. The analysis reveals that synthetic data can effectively replace original data for training two commonly used machine learning models, random forest and linear regression, yielding low MAPE errors on actual test data.

\begin{table}[h]
    \small
    \centering
    \caption{MLU- Random Forest and Linear Regression.}
    \begin{tabular}{l|c|cc}
        \toprule
        \multicolumn{4}{c}{Machine Learning Utility (MLU) - Random Forest ($\downarrow$)~} \\
        \midrule
        Dataset & \makecell{W/o\\KGP} & \makecell{Statistical\\KGP} & \makecell{Semantic\\KGP} \\
        \midrule
        AP Calculus & 0.47$\pm$0.45 & 0.30$\pm$0.32 & \textbf{0.27$\pm$0.31} \\
        \midrule
        Datasaurus Dozen & 0.90$\pm$0.30 & 0.88$\pm$0.25 & \textbf{0.60}$\pm$\textbf{0.19}  \\
        \midrule
         $O_2$ Sensing & 0.031 & 0.029 & \textbf{0.0225} \\
        \bottomrule
        \toprule
        \multicolumn{4}{c}{Machine Learning Utility (MLU) - Linear Regression ($\downarrow$)~} \\
        \midrule
        Dataset & \makecell{W/o\\KGP} & \makecell{Statistical\\KGP} & \makecell{Semantic\\KGP} \\
        \midrule
        AP Calculus &1.61$\pm$2.02 & 1.61$\pm$2.02& 2.13$\pm$2.90\\
        \midrule
        Datasaurus Dozen & 0.90$\pm$0.13 & 0.84$\pm$0.08 & \textbf{0.79}$\pm$\textbf{0.08} \\
        \midrule
         $O_2$ Sensing & 0.039 & 0.053& \textbf{0.023} \\
        \bottomrule
    \end{tabular}
    \label{tab:mle}
\end{table}

\par\noindent
\textit{\ul{Overall Finding}}: \textit{Using KGP (Statistical and Semantic) resulted in optimal performance across all three standard synthetic table metrics, with an average enhancement of 50\% for each metric.}

\subsection{Case Study: Characterizing Effectiveness of Prompt Optimization via KGP in a Real-World Cyber-Physical Scenario}
\label{sec:case1}
 
This section uses a dataset from a noisy cyber-physical system recording temperature, salinity, and pressure to predict water's oxygen solubility. 
Table~\ref{tab:dcr_noised} presents the evaluation of generating synthetic data from noisy raw data using Statistical KGP and Semantic KGP.

\begin{table}[!htbp]
\small
\centering%
\caption{Distance to the closest record.}
\begin{tabular}{lccc}
\toprule
\multicolumn{4}{c}{Distance to the closest record under Noisy Case. ($\downarrow$)~} \\
\midrule
Dataset & W/o KGP & Statistical KGP & Semantic KGP \\
\midrule
$O_2$ Sensing W/o Noise & 0.57 & 0.46 & \textbf{0.38}  \\
\midrule
$O_2$ Sensing W/ Noise &1.06 & \textbf{0.61}& 0.70\\
\bottomrule
\end{tabular}
\label{tab:dcr_noised}
\end{table}

\par\noindent
\textit{\ul{Finding}}: \textit{Statistical KGP is vital for preserving a valid row joint distribution in scenarios characterized by noisy data.}

One of the advantages of using modern LLMs is the availability and flexibility of the agent-embedded framework.
Considering that even explicitly including the instruction ``does not copy the original data'' in the prompt, the generated data may still include some, see Table~\ref{tab:dcr_table}.
Given that the LLM generator can utilize foundational and supplementary domain knowledge (statistical and semantic KGPs) to correct errors, we will initially introduce noise to the original data and subsequently employ the LLM to correct this noise, a process referred to as \textbf{ noise-and-refix}. 
When the real original data are not presented to the LLM agent, concerns regarding the copying or leaking of data are eliminated.

\par\noindent
\textit{\ul{Finding}}: \textit{A modicum of in-context example data is essential to generate the hidden distribution, while knowledge guidance is more effective in providing dependency, correcting errors, or establishing boundaries.}

\section{Conclusion}

This paper proposes the use of a novel prompt optimization strategy termed Knowledge-Guided Prompting (KGP), to enhance the generation quality of structural tabular data by a Large Language Model (LLM). Although examples of in-context learning data are limited in a chunk size and offer localized knowledge only, the comprehensive domain knowledge of the entire dataset can be incorporated as an English prompt using statistical KGP and semantic KGP.
We have investigated the relationship between symbolic and statistical knowledge and prompt snippets, yielding an empirical `scaling law' that estimates the number of snippets needed.
Our experiments demonstrate that the KGP strategy can reduce ICL data (that is, tokens) by 40\%, improve data with unknown regions (out-of-distribution generation) or improve the quality of synthetic data while utilizing the same level of ICL data. 

Future work will be aimed at developing a multimodal learning framework encompassing KGP with visual and semantic facets. 
A user's inconsistent semantic KGP, when compared to the example data, may result in a decrease in generation quality, constituting a form of model poisoning. Leveraging associations across modalities via shared
parameters will lead to more
resilient approaches for knowledge-guided
applications.

\clearpage

\clearpage
\bibliographystyle{ACM-Reference-Format}
\bibliography{sections/reference}

\end{document}